\documentclass[runningheads]{llncs}

\usepackage{eccv}

\usepackage{eccvabbrv}

\usepackage{graphicx}
\usepackage{booktabs}

\usepackage[accsupp]{axessibility}  

\usepackage{hyperref}

\usepackage{orcidlink}

\usepackage{multirow}

\begin{document}

\title{LensStyle: Learning the Aesthetics of Lenses for Controllable Optical Style Rendering} 

\titlerunning{LensStyle}

\author{Yachuan Huang \orcidlink{0009-0004-2314-3992} \and
Liwen Xiao \orcidlink{0009-0002-7739-870X} \and
Liao Shen \orcidlink{0000-0002-2423-4835} \and
Qiwen Wang \orcidlink{0009-0007-4520-0478} \and Huiqiang Sun \orcidlink{0000-0002-3653-3613}\thanks{Corresponding author.} \and Zhiyu Pan \orcidlink{0000-0001-5584-6669} \and Zhiguo Cao \orcidlink{0000-0002-9223-1863}}
\authorrunning{Y.~Huang et al.}

\institute{School of AIA, Huazhong University of Science and Technology, Wuhan, China
\email{\{yachuanhuang,liwenxiao,leoshen,qiwenwang,shq1031,zhiyupan,zgcao\}@hust.edu.cn}}
\maketitle

\begin{abstract}
    The visual aesthetics of photographs are deeply influenced by lens characteristics such as aperture shape, optical vignetting and optical diffraction, which together define a camera’s unique optical style.
    Existing lens effect rendering methods primarily focus on accurately simulating the blur transition from small to large apertures but overlook the stylistic aspects of lens effects.
    As a result, they fail to produce diverse bokeh effects under large apertures or capture distinctive photographic phenomena such as starbursts that emerge under small apertures.
    In this work, we introduce LensStyle, a unified framework for controllable stylized lens effect rendering that explicitly models lens aesthetics through joint continuous–discrete control.
    Our model incorporates a Dual-Path Controller that disentangles continuous optical parameter modulation (\eg, focus distance and blur strength) from discrete lens-style conditioning (\eg, circular, polygonal, donut, cat-eye, and starburst effects), enabling fine-grained, interpretable, and physically grounded lens manipulation within a single unified framework.
    To support model training, we curate a comprehensive MultiLens dataset containing multi-lens image pairs synthesized under real optical constraints.
    Extensive experiments demonstrate that LensStyle achieves superior realism, controllability, and aesthetic quality compared with existing lens effect rendering approaches and diffusion-based image editing models, advancing computational photography toward multiple-lens-style simulation.
\end{abstract}

\begin{figure}[!t]
	\centering
	\includegraphics[width=\textwidth]{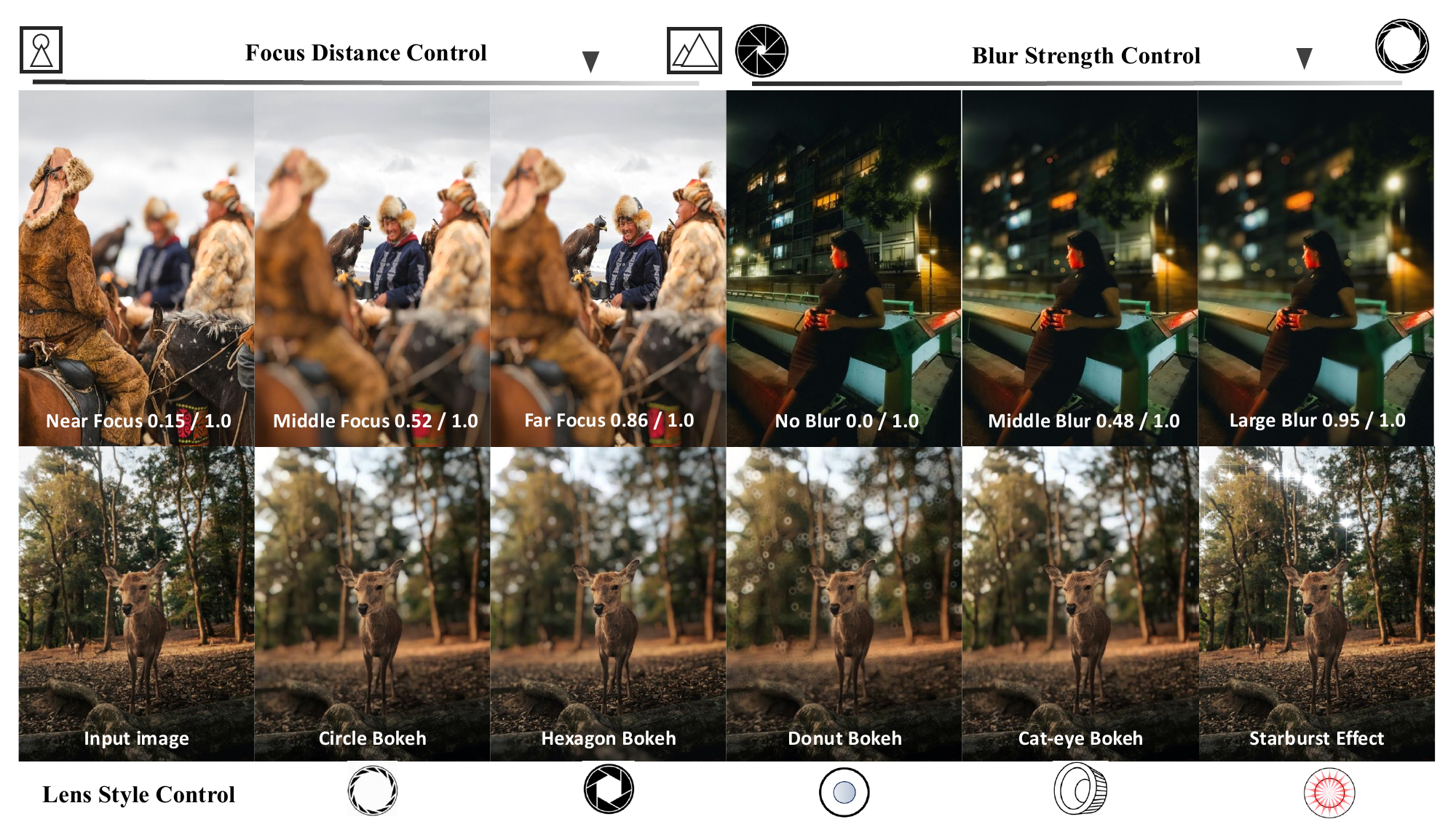}
    \caption{\textbf{Controllable multi-style lens effect rendering with LensStyle.}
    Our framework enables unified control over three complementary dimensions of lens aesthetics within a single model. 
    \emph{Top-left:} continuous focus distance control, where the focal plane smoothly shifts from near to far. 
    \emph{Top-right:} continuous blur strength control, adjusting the magnitude of defocus under fixed focus. 
    \emph{Bottom:} discrete lens style control, synthesizing diverse aperture-induced effects including circular, hexagonal, donut, cat-eye bokeh, and diffraction-driven starburst patterns. 
    These results demonstrate that LensStyle jointly models continuous optical parameters and discrete lens geometries in a physically grounded and interpretable manner.}
	\label{fig:figure1}
\end{figure}

\section{Introduction}
\label{sec:intro}

The visual aesthetics of a photograph are profoundly influenced by the optical characteristics of the camera lenses.
Factors such as aperture geometry, optical vignetting, and diffraction patterns determine how light is transmitted and scattered, collectively defining a camera’s unique lens style.
Different optical designs lead to distinctive perceptual outcomes.
A circular or polygonal aperture produces distinctive bokeh shapes under large apertures; spherical aberration or mirror-based lens designs give rise to donut-shaped bokeh with hollow highlights; optical vignetting along the image periphery introduces the characteristic cat-eye effect; and small-aperture diffraction generates starburst patterns radiating from bright point sources.
Such phenomena not only enhance image realism but also constitute the artistic signature of individual lenses and camera systems, playing a vital role in photographic aesthetics and visual perception.

Recent advances in lens effect rendering have primarily focused on bokeh synthesis.
Methods such as classical and neural bokeh rendering~\cite{yang2016virtual, wadhwa2018synthetic, sheng2024dr,wang2018deeplens,peng2022bokehme,luo2023defocus,
seizinger2025bokehlicious} or diffusion-driven bokeh generation~\cite{yuan2025generative, fortes2025bokeh, zhu2025bokehdiff, qincamedit, huang2026bokehflow} can simulate the blur transition from small to large apertures, enabling synthetic shallow depth-of-field from all-in-focus inputs, typically with uniform, circular bokeh patterns.
In other words, these approaches primarily model the \textit{amount} of blur while neglecting its \textit{style}.
In practice, the appearance of bokeh is not only uniform, it also depends on the number of aperture blades, the shape of the aperture and optical aberrations, which jointly determine the texture, shape, and highlight distribution within the defocus region.
Moreover, images captured with a small aperture often exhibit aesthetically pleasing starburst effect~\cite{liu2016star}, where radiating diffraction spikes emerge from light diffraction at the aperture edges—an effect that existing lens rendering approaches have yet to address.

To bridge the gap, we introduce \textbf{LensStyle}, a unified flow-based framework for controllable lens effect rendering that enables joint control over continuous optical parameters and diverse lens styles, explicitly modeling the aesthetic characteristics of real-world lenses.
Unlike prior works that produce only a single bokeh style, LensStyle models lens behavior as a learnable style space, capturing diverse optical phenomena ranging from circular, polygonal, donut and cat-eye bokeh to diffraction-induced starbursts. 
At the core of our framework lies a \textbf{Dual Path Controller}, which disentangles continuous optical control from discrete lens-style modulation. 
For continuous attributes such as focus distance and blur strength, we introduce a \textit{Continuous Optical Parameter Modulation} branch. 
User-specified optical parameters are first encoded through an optics adapter into optical parameter embeddings, which are then transformed into adaptive scaling and shifting factors via lightweight MLPs. 
These factors modulate intermediate image features in the flow matching network, enabling precise and smooth control over focal plane and defocus magnitude adjustment.
In parallel, discrete lens-style categories are handled by a \textit{Discrete Lens Style Cross-Attention} branch. 
The selected lens style is encoded into text embeddings and injected into the generative backbone via cross-attention, guiding the model to synthesize style-specific optical characteristics such as aperture geometry, vignetting-induced cat-eye distortion, or diffraction-driven starbursts.
By decoupling continuous optical parameter modulation from discrete style conditioning, LensStyle achieves fine-grained, interpretable, and unified control over a wide spectrum of lens effects within a single flow-based generative model.

To enable comprehensive lens modeling, we present the MultiLens dataset, a large-scale benchmark containing physically grounded multi-lens synthetic pairs with optical parameters (focus distance and blur strength). As the first dataset designed for unified lens-style rendering, MultiLens provides structured supervision for diverse optical effects, including circular, polygonal, donut, and cat-eye bokeh, along with diffraction-induced starbursts.

Our contributions can be summarized as follows:

\begin{itemize}
\item We propose \textbf{LensStyle}, a unified flow-based framework for controllable lens effect rendering that jointly models continuous optical parameters and discrete lens-style categories within a single generative architecture.

\item We introduce a \textbf{Dual Path Controller}, which disentangles continuous optical parameter modulation from discrete lens-style conditioning, enabling fine-grained, interpretable, and photorealistic control over focus distance, blur strength, and diverse optical styles.

\item We construct the \textbf{MultiLens} dataset, a large-scale, physically grounded lens effect benchmark that offers continuous control over focus distance and blur strength. More importantly, it spans multiple lens styles, covering circular, polygonal, donut, cat-eye bokeh, and diffraction-induced starburst effects.
\end{itemize}

\section{Related Work}
\label{sec:related_work}
\subsection{Lens Effect Rendering}
Existing lens effect rendering methods mainly focus on generating shallow depth-of-field bokeh images from all-in-focus inputs.
Early traditional approaches~\cite{yang2016virtual, wadhwa2018synthetic, zhang2019synthetic} rely on depth maps and spatially-varying blur kernels to approximate large-aperture effects.
Subsequent work, such as Dr.Bokeh~\cite{sheng2024dr}, introduces an improved compositing formulation that alleviates artifacts under complex occlusion conditions. With the advent of deep learning, depth-dependant neural rendering approaches~\cite{xiao2018deepfocus,wang2018deeplens,peng2022bokehme,peng2022mpib,luo2023defocus, luo2024video} leverage depth maps and neural networks to produce more physics-aware bokeh effects, while depth-free method bokehlicious~\cite{seizinger2025bokehlicious} synthesize bokeh patterns based on a large-scale camera-captured paired dataset. More recently generative models~\cite{zhu2025bokehdiff, qincamedit, diffcamera,shrivastava2025defocusblur,huang2026bokehflow,voynov2024curved} further advance the field by leveraging the strong priors from diffusion or flow-based generators and design controllable architectures to achieve bokeh rendering.
Beyond general bokeh rendering, several works have explored depth-of-field extension~\cite{luo2022point} or starburst simulation~\cite{liu2016star}.
However, existing bokeh rendering methods can only reproduce a single type of bokeh pattern, typically the standard circular bokeh or the dataset-specific style, while neglecting the stylistic and aesthetic characteristics of real lenses.
In this paper, we propose a unified framework that generates diverse lens-style images and controls continuous focus distance and blur strength.
\subsection{Generative Models}
Recent diffusion models have demonstrated remarkable success in both unconditional image synthesis~\cite{saharia2022photorealistic, ramesh2021zero, song2020score,nichol2021improved} and controllable image generation~\cite{zhang2023adding, gal2022image,ruiz2023dreambooth}.
Representative works such as DDPM~\cite{ho2020denoising} and Stable Diffusion~\cite{rombach2022high} enable high-quality generation from textual prompts through iterative denoising.
Building on these, numerous diffusion-based editing approaches have emerged, including InstructPix2Pix~\cite{brooks2023instructpix2pix}, SDEdit~\cite{meng2021sdedit}, and SuperEdit~\cite{li2025superedit}, which support fine-grained semantic modification through conditional diffusion steps or prompt-based control.
In parallel, Flow matching~\cite{lipman2022flow,dao2023flow,flux2024} and rectified flow~\cite{liu2022flow,albergo2022building} models provide an alternative paradigm to diffusion by learning deterministic mappings between data and noise distributions.
Unlike diffusion models that rely on iterative denoising, flow-based models directly predict continuous-time velocity fields, enabling highly efficient one-step or few-step sampling with significantly higher efficiency~\cite{meanflow,geng2025improved,salimans2022progressive, sauer2024adversarial}.
Recent works~\cite{kulikov2025flowedit,hulfm,zhao2024ultraedit} extend this formulation to image editing tasks, achieving competitive quality with much lower computational cost.
In this work, we adopt a flow-based generative formulation as the foundation of our LensStyle framework.
This design allows us to model lens-style rendering as a continuous transport from an all-in-focus image to a stylistically one, efficiently capturing both spatial coherence and optical consistency.

\begin{figure*}[!t]
	\centering
	\includegraphics[width=\linewidth]{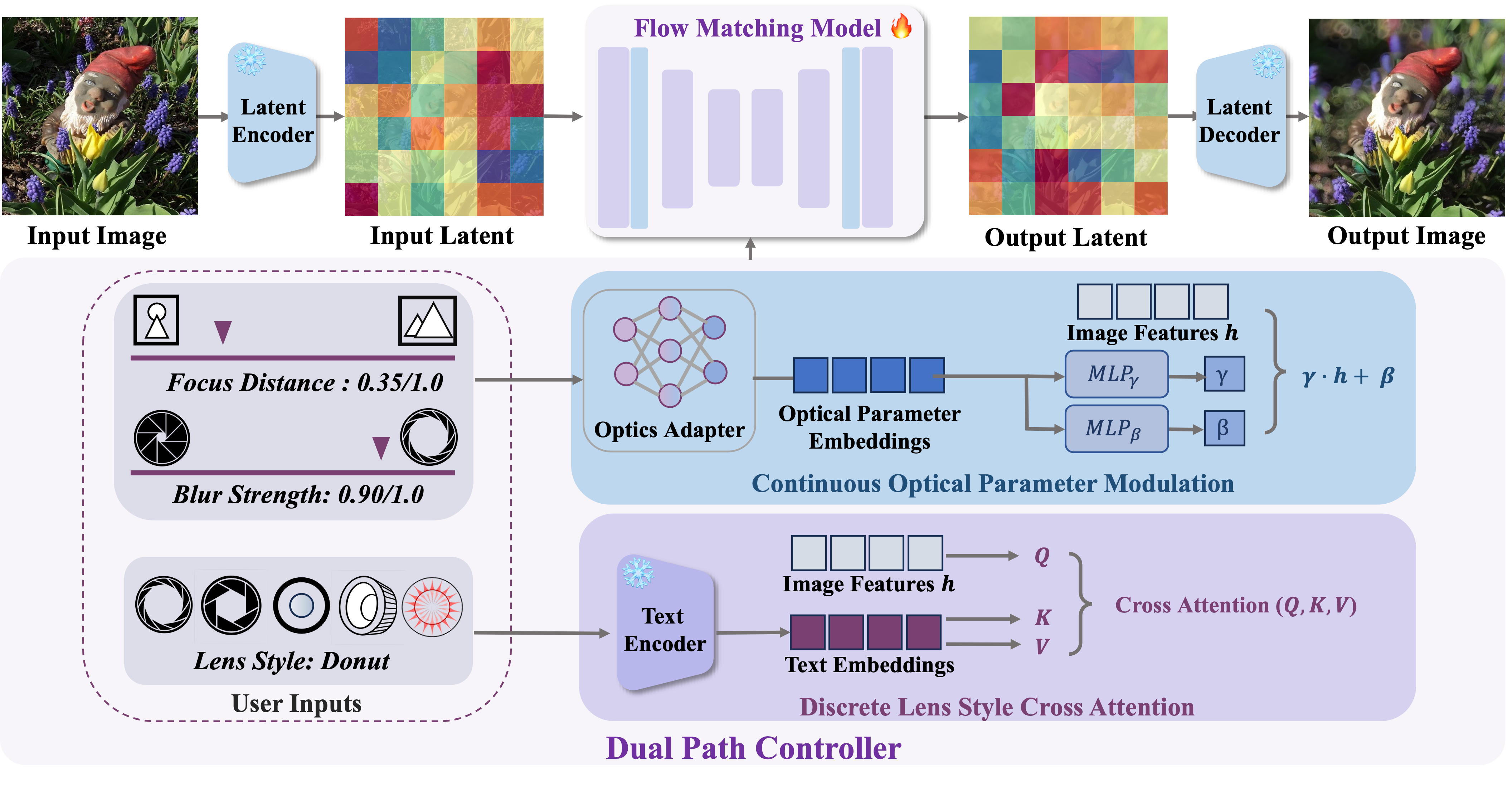}
	\caption{\textbf{Overview of LensStyle.}
    The input image is first encoded into latent by a VAE encoder and then processed by a flow matching network for lens effect rendering. 
    A Dual Path Controller conditions the generative backbone through two complementary branches. 
    The \emph{Continuous Optical Parameter Modulation} branch encodes user-specified optical parameters (\eg, focus distance and blur strength) into embeddings that generate adaptive scaling and shifting factors, which modulate intermediate image features for precise continuous control. 
    In parallel, the \emph{Discrete Lens Style Cross-Attention} branch encodes the selected lens style into text embeddings and injects style-specific information into the network via cross-attention. 
    The denoised latent is finally decoded by the VAE decoder to produce the stylized output image.}
	\label{fig:pipeline}
\end{figure*}
\section{Method}
\label{sec:method}

\subsection{LensStyle Framework}
\label{subsec:lensstyle}
We aim to model the \textit{aesthetic qualities of lenses}, which arise from underlying optical properties such as aperture geometry, focal configuration, vignetting, and diffraction.

As illustrated in Figure~\ref{fig:pipeline}, our proposed \textbf{LensStyle} adopts a flow-based generative architecture composed of three key components: 
(1) a \textit{latent autoencoder} to encode all-in-focus images into a perceptually aligned latent space and a \textit{latent decoder} to reconstruct images from the latent space;
(2) a \textit{Dual Path Controller} that enables joint continuous–discrete optical control;
and (3) a \textit{flow-based rendering model} that transforms latent representations of all-in-focus images into corresponding 
styled ones.

Unlike prior lens rendering approaches that explicitly require disparity or depth maps as additional inputs~\cite{peng2022bokehme,sheng2024dr,peng2022mpib,zhu2025bokehdiff}, 
LensStyle operates in a depth-free manner ~\cite{seizinger2025bokehlicious, qincamedit} during training and inference. 
This avoids dependence on potentially unreliable depth estimation in challenging real-world scenarios.
Depth is used only during dataset synthesis (Section~\ref{subsec:dataset}) to ensure physically consistent supervision. 
Through learning depth-consistent blur transitions during training, the model implicitly internalizes scene geometry within its latent representation, enabling continuous focus and blur control without explicit depth conditioning.

Let $I_A$ and $I_L$ denote the all-in-focus and lens-style images, respectively. 
We first obtain latent codes $z_A = \mathcal{E}(I_A)$ and $z_L = \mathcal{E}(I_L)$ using an encoder $\mathcal{E}$. 

Inspired by the success of the flow matching paradigm in image generation~\cite{dao2023flow,liu2022flow,lipman2022flow,fischer2023boosting}, we directly learn a deterministic vector field $N_{fm}$ that maps $z_A$ to $z_L$ in our method:
\begin{equation}
\mathcal{L}_{FM} = \mathbb{E}_t \| N_{fm}(\phi_t(z_L), z_A, z_C, z_D, t) -v_t(\phi_1(z_L))  \|_2^2,
\label{eq:lfm}
\end{equation}
where $\phi_t(z_L) = (1-t)z_A + t z_L$ defines the linear interpolation between input $ z_A $ and target $ z_L $, 
$z_C$ denotes continuous optical parameter embedding, and $z_D$ is the discrete lens style conditioning embedding from the Dual Path Controller.
The objective $v_t(\phi_1(z_L)) = \phi_1(z_L) - \phi_t(z_L)$ enables the flow matching network to directly predict the residual vector for linear transport from the current distribution to the target distribution, at varying scales.



\subsection{Dual Path Controller}
\label{subsec:lenscontroller}

The Dual Path Controller enables unified yet disentangled control over 
(i) continuous optical parameters and 
(ii) discrete lens-style categories.
Instead of treating lens rendering as purely semantic editing, 
we explicitly separate physically grounded parameter modulation 
from style-specific feature conditioning.

\subsubsection{Continuous Optical Parameter Modulation.}

Continuous optical attributes, such as focus distance and blur strength, are encoded into optical parameter embeddings $\mathbf{p}$ via an optics adapter. Specifically, the adapter first normalizes the user-specified focus distance and blur magnitude, and then applies Fourier feature encoding to transform them into a high-dimensional embedding vector $\mathbf{p}$.

Given an embedding $\mathbf{p}$, 
we generate adaptive scaling and shifting coefficients 
$\gamma(\mathbf{p})$ and $\beta(\mathbf{p})$ 
through lightweight MLPs, and modulate intermediate features $h$ of the flow backbone as:

\begin{equation}
h' = \gamma(\mathbf{p}) \odot h + \beta(\mathbf{p}).
\end{equation}

This mechanism enables smooth and interpretable control over defocus magnitude and focal plane adjustment, 
while preserving spatial coherence of the input image.

\subsubsection{Discrete Lens Style Cross-Attention.}

Discrete lens-style categories (e.g., circular, polygonal, donut, cat-eye, starburst) 
are encoded into style embeddings $z_T$ using a text encoder.

These embeddings are injected into the flow backbone through cross-attention:

\begin{equation}
\mathrm{CrossAttn}(Q,K,V) =
\mathrm{softmax}\!\left(\frac{QK^\top}{\sqrt{d}}\right)V,
\end{equation}

with $Q = W_Q h'$, $K = W_K z_T$, and $V = W_V z_T$.

This branch governs high-level stylistic characteristics such as aperture geometry,
vignetting-induced distortion, and diffraction spike patterns,
ensuring consistent rendering of discrete lens aesthetics.

By disentangling continuous optical modulation from discrete style conditioning,
the Dual Path Controller enables fine-grained, interpretable,
and physically grounded lens effect rendering within a single generative model.

\subsection{MultiLens Dataset Synthesis}
\label{subsec:dataset}

To enable learning of diverse and physically grounded optical styles, we construct the \textbf{MultiLens} dataset using a unified optical forward model that jointly simulates defocus blur and diffraction effects.

The all-in-focus images $I_A$ are sourced from ReDWeb~\cite{xian2018monocular}, a real-world monocular depth benchmark that provides ground-truth depth maps $Z(p)$.
The availability of accurate depth enables physically consistent synthesis of spatially varying defocus, producing realistic blur transitions during training.
Notably, depth is used only for data synthesis. At inference time, our model does not require depth as input, making depth-free lens effect rendering feasible through physically grounded supervision.

\subsubsection{Unified Optical Forward Model.}
Let $A(x,y)$ denote the aperture transmission function defined on the aperture plane, where $(x,y)$ are aperture-plane coordinates and $A(x,y)\in\{0,1\}$ indicates whether light passes through the aperture at location $(x,y)$.
Under the thin-lens approximation, image formation is modeled as the combination of (i) defocus blur described by geometric optics and (ii) diffraction effects described by wave optics.

\begin{figure}[tb]
  \centering
  \begin{subfigure}{0.46\linewidth}
    \includegraphics[width=\linewidth]{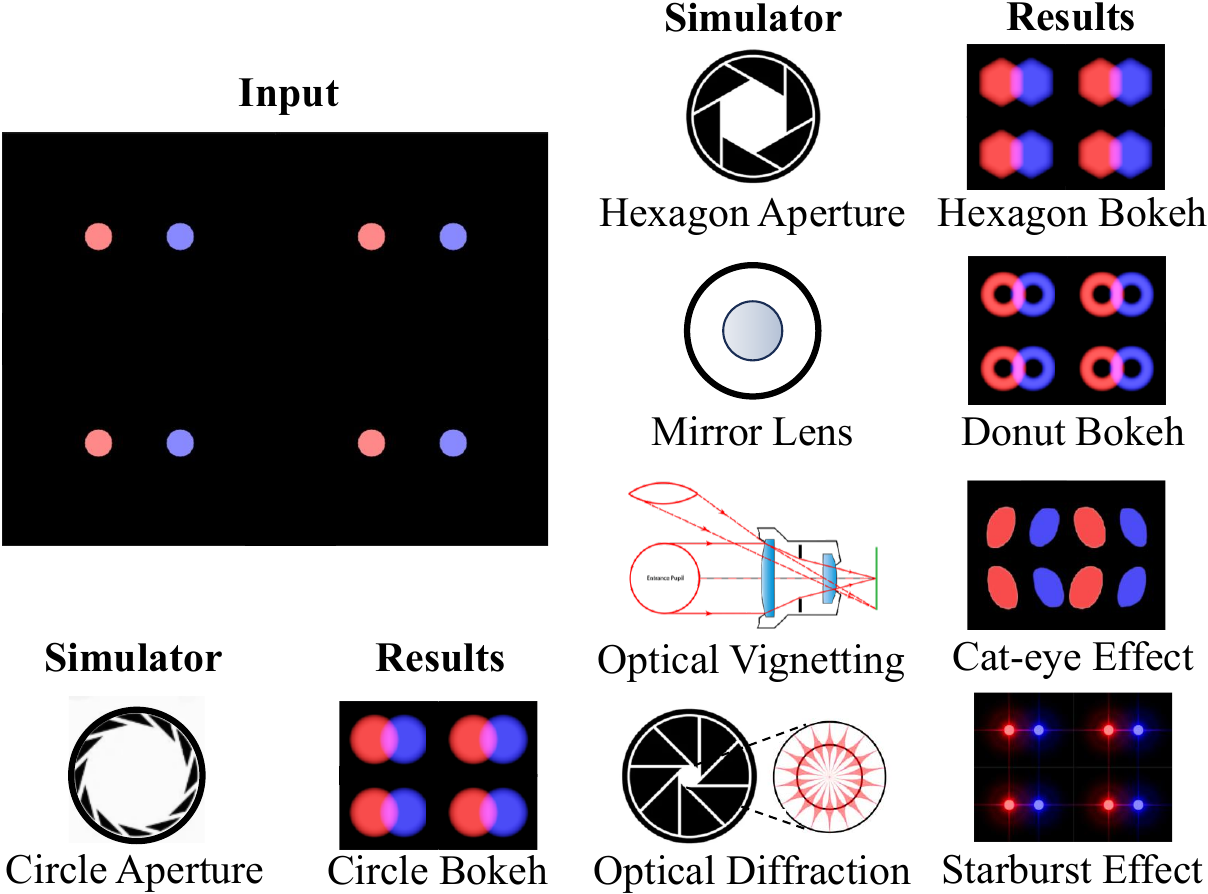}
    \caption{\textbf{Illustration of MultiLens Synthesis.} Our simulator maps point inputs to diverse optical effects, including circle bokeh, hexagonal bokeh, donut bokeh, cat-eye effects from optical vignetting, and starbursts from diffraction.
    }
    \label{fig:dataset-a}
  \end{subfigure}
  \hfill
  \begin{subfigure}{0.50\linewidth}
    \includegraphics[width=\linewidth]{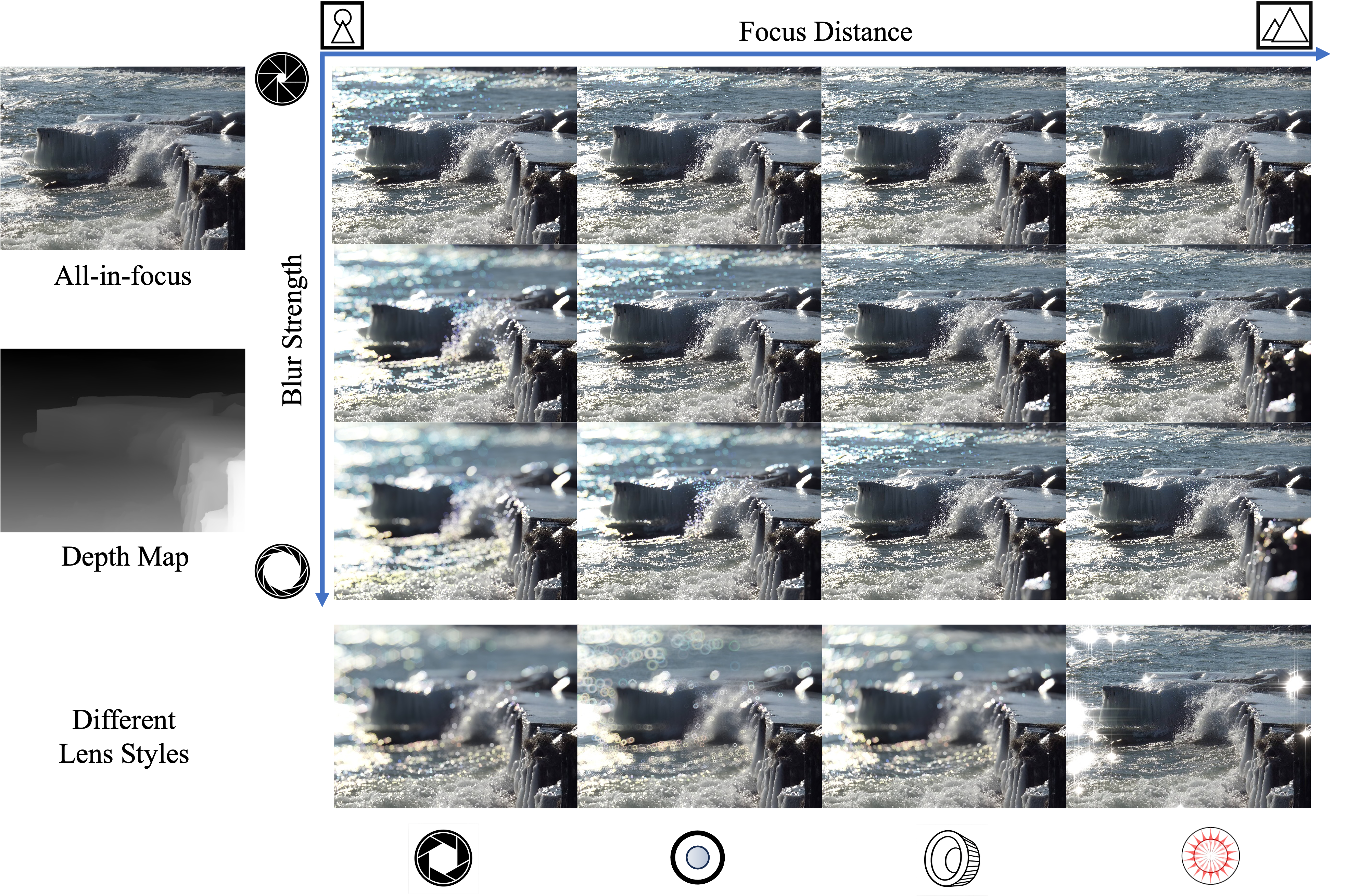}
    \caption{\textbf{Examples of the MultiLens dataset.} 
   A sample grid generated from an all-in-focus image and depth map, demonstrating continuous control over focus distance (x-axis) and blur strength (y-axis) and different lens style configurations.
    }
    \label{fig:dataset-b}
  \end{subfigure}
  \caption{MultiLens Dataset Overview}
  \label{fig:dataset}
\end{figure}

\paragraph{Defocus PSF.}
The defocus point spread function (PSF) is proportional to the projected aperture:
\begin{equation}
B(x,y)=\frac{A(x,y)}{\sum_{x',y'}A(x',y')},
\label{eq:psf_defocus}
\end{equation}
where $B(x,y)$ is a normalized blur kernel satisfying energy conservation.

\paragraph{Diffraction PSF.}
To synthesize starburst effects under small apertures, we adopt Fraunhofer diffraction:
\begin{equation}
D(u,v)=\left|\mathcal{F}\{A(x,y)\}\right|^2,
\label{eq:psf_diffraction}
\end{equation}
where $\mathcal{F}$ denotes the 2D Fourier transform and $(u,v)$ are spatial-frequency coordinates proportional to angular deviations on the sensor plane.
This formulation yields diffraction spikes whose count and orientation are determined by the aperture blade number $N_b$ and rotation $\theta_0$.

\subsubsection{Aperture Functions for Multi-style Bokeh.}
Different bokeh styles are modeled by different analytical forms of the aperture transmission $A(x,y)$.

\paragraph{(1) Circular aperture.}
For a circular aperture with radius $r_0$:
\begin{equation}
A_{\mathrm{circ}}(x,y)=
\begin{cases}
1, & x^2+y^2\le r_0^2,\\
0, & \text{otherwise.}
\end{cases}
\label{eq:aperture_circ}
\end{equation}

\paragraph{(2) Polygonal aperture.}
For a regular $N_b$-sided polygon aperture rotated by $\theta_0$:
\begin{equation}
A_{\mathrm{poly}}(x,y)=
\begin{cases}
1, & \forall k,\; x\cos\theta_k+y\sin\theta_k \le r_0\cos\!\left(\frac{\pi}{N_b}\right),\\
0, & \text{otherwise,}
\end{cases}
\label{eq:aperture_poly}
\end{equation}
where $\theta_k=\frac{2\pi k}{N_b}+\theta_0$ specifies the orientation of each blade edge.

\paragraph{(3) Donut aperture.}
To model mirror-lens or strong spherical-aberration cases, we use a ring-shaped aperture:
\begin{equation}
A_{\mathrm{donut}}(x,y)=
\begin{cases}
1, & r_1^2 \le x^2+y^2 \le r_2^2,\\
0, & \text{otherwise,}
\end{cases}
\label{eq:aperture_donut}
\end{equation}
where $r_2$ is the outer radius and $r_1=\mu r_2$ controls the inner void with $\mu\in[0,1)$.

\paragraph{(4) Cat-eye aperture.}
The cat-eye effect caused by vignetting under off-axis viewing is approximated by an elliptically clipped aperture:
\begin{equation}
A_{\mathrm{cat}}(x,y)=
\begin{cases}
1, & \frac{(x-\delta_x)^2}{a^2}+\frac{y^2}{b^2}\le 1,\\
0, & \text{otherwise,}
\end{cases}
\label{eq:aperture_cat}
\end{equation}
where $\delta_x$ models the off-axis shift, and $a,b$ are ellipse axes with $b/a<1$ controlling lateral compression.

\subsubsection{Rendering with Depth-consistent Defocus and Diffraction.}
Given $I_A$ and depth map $Z(p)$, we define a spatially varying defocus magnitude:
\begin{equation}
K(p)=\alpha\,|Z(p)-Z_f|,
\label{eq:defocus_strength}
\end{equation}
where $Z_f$ is the focal depth and $\alpha$ controls blur strength.
The final stylized image is synthesized by spatially varying convolution with a kernel scaled by $K(p)$:
\begin{equation}
I_L(p)=\sum_q I_A(q)\cdot B_{K(p)}(p-q) \;+\; \lambda_d \cdot H(I_A)(p)\cdot D(p),
\label{eq:render_full}
\end{equation}
where $B_{K(p)}$ denotes the defocus PSF whose effective radius is proportional to $K(p)$,
$H(I_A)$ extracts high-intensity regions (e.g., saturated highlights) to trigger visible diffraction spikes,
$D(p)$ is the diffraction response projected onto the image plane, and $\lambda_d$ controls diffraction strength.

\subsubsection{Dataset Composition.}

MultiLens comprises 303,000 paired samples derived from 3,000 all-in-focus scenes. The dataset covers (1) 300,000 defocus pairs, generated by varying 4 aperture geometries $\{A_{\mathrm{circ}},A_{\mathrm{poly}},A_{\mathrm{donut}},A_{\mathrm{cat}}\}$ across 5 focus distances $Z_f$ and 5 blur strengths $\alpha$; and (2) 3,000 diffraction pairs capturing starburst effects by modulating the diffraction parameter $\lambda_d$. This structured composition ensures a comprehensive coverage of physically grounded lens effects with precisely controlled optical parameters.
\subsection{Optimization.}
\label{subsec:supervision}

We supervise the model with a combination of flow consistency and perceptual reconstruction losses.  
The overall objective is:
\begin{equation}
\mathcal{L} = \mathcal{L}_{FM} + \lambda_1 \mathcal{L}_{rec} 
+ \lambda_2 \mathcal{L}_{perc} ,
\label{eq:loss}
\end{equation}
where $\mathcal{L}_{FM}$ is the flow matching loss described in Eq.~\eqref{eq:lfm}, 
$\mathcal{L}_{rec} = \| D(\hat{z}_L) - I_L \|_1$ enforces pixel-level fidelity via the decoder $D$, $\mathcal{L}_{perc}$ is a perceptual loss computed using a pretrained VGG network~\cite{simonyan2015vgg} to align high-level aesthetics.

\section{Experiments}
\label{sec:experiments}

\subsection{Experimental Settings}
\subsubsection{Baselines and Metrics.} For the standard circular bokeh, we adopt several open-source state-of-the-art models as baselines, including MPIB~\cite{peng2022mpib}, BokehMe~\cite{peng2022bokehme}, Dr.Bokeh~\cite{sheng2024dr}, and BokehDiff~\cite{zhu2025bokehdiff}.
Since our work is the first to explicitly consider bokeh style, no existing bokeh rendering methods can directly generate polygonal, donut, or cat-eye bokeh effects.
Therefore, we further compare our approach with state-of-the-art instruction-based image editing methods, including UltraEdit~\cite{zhao2024ultraedit} and SuperEdit~\cite{li2025superedit}. As they do not take lens parameters into consideration, we further retrain SuperEdit and UltraEdit on our MultiLens dataset to enable lens style editing, denoted as SuperEdit* and UltraEdit*, for fair comparison. Training details can be found in our appendix. For the starburst effect, although prior work \cite{liu2016star} has explored the physical simulation of starbursts, it has not released its implementation. Hence, we also employ image editing models for comparison in this setting. Following prior work~\cite{peng2022bokehme,zhu2025bokehdiff}, we report the PSNR and SSIM scores. Since PSNR is insensitive to blurring \cite{sheng2024dr}, we additionally include LPIPS \cite{zhang2018unreasonable}, a perceptual metric that better aligns with human visual perception.
\subsubsection{Implementation Details.}

Training is performed on the MultiLens dataset at a resolution of $512\times384$ for 10K iterations using the Adam optimizer~\cite{kingma2014adam}. 
The initial learning rate is set to $3\times10^{-5}$ and linearly decayed to $3\times10^{-7}$ after 3K iterations.
We adopt an 8$\times$ downsampling VAE~\cite{kingma2013auto} that produces a 4-channel latent representation. 
The flow-based backbone is implemented as a conditional U-Net with 8 input channels and 4 output channels, and is initiated from pretrained SD2.1~\cite{rombach2022high} using the Diffusers library.
In the Continuous Optical Parameter Modulation branch, both $MLP_\gamma$ and $MLP_\beta$ are implemented as 3-layer multilayer perceptrons with SiLU activations. 
For the Discrete Lens Style Cross-Attention branch, we employ a pretrained CLIP text encoder~\cite{radford2021learning} to extract style embeddings.
The loss coefficients in Eq.~\eqref{eq:loss} are set to $\lambda_1=0.2$ and $\lambda_2=0.5$.
For diffusion-based UltraEdit and SuperEdit, the number of inference steps is set to 50, with the image guidance scale and text guidance scale configured to 1.5 and 7.5, respectively.
All experiments are conducted on the NVIDIA RTX A6000 GPU.

\subsection{Results and Comparisons}
\begin{table*}[t]
\centering

\caption{\textbf{Quantitative comparison across polygonal, donut, cat-eye, and starburst styles on MultiLens dataset.} Higher PSNR/SSIM and lower LPIPS are better. LensStyle constantly outperforms diffusion-based image editing methods.}
\resizebox{\textwidth}{!}{
\begin{tabular}{l ccc ccc ccc ccc}
\toprule
\multirow{2}{*}{\textbf{Method}} &
\multicolumn{3}{c}{\textbf{Hexagon}} &
\multicolumn{3}{c}{\textbf{Donut}} &
\multicolumn{3}{c}{\textbf{Cat-eye}} &
\multicolumn{3}{c}{\textbf{Starburst}} \\
\cmidrule(lr){2-4} \cmidrule(lr){5-7} \cmidrule(lr){8-10} \cmidrule(lr){11-13}
 & PSNR $\uparrow$ & SSIM $\uparrow$ & LPIPS $\downarrow$
 & PSNR $\uparrow$ & SSIM $\uparrow$ & LPIPS $\downarrow$
 & PSNR $\uparrow$ & SSIM $\uparrow$ & LPIPS $\downarrow$
 & PSNR $\uparrow$ & SSIM $\uparrow$ & LPIPS $\downarrow$ \\
\midrule
UltraEdit~\cite{zhao2024ultraedit}  & 14.57 & 0.3808 & 0.5971 & 14.52 & 0.3845 & 0.5504 & 14.43 & 0.3790 & 0.5833 & 14.32 & 0.3785 & 0.5821 \\
SuperEdit~\cite{li2025superedit}    & 17.27 & 0.4900 & 0.4320 & 16.24 & 0.4608 & 0.4180 & 17.58 & 0.4981 & 0.4043 & 17.42 & 0.4975 & 0.4021 \\
UltraEdit*~\cite{zhao2024ultraedit} & 19.11 & 0.6109 & 0.3240 & 19.21 & 0.5977 & 0.2811 & 18.60 & 0.5738 & 0.3436 & 18.67 & 0.5725 & 0.3445 \\
SuperEdit*~\cite{li2025superedit}   & \underline{22.77} & \underline{0.7230} & \underline{0.1835} & \underline{22.27} & \underline{0.6723} & \underline{0.1871} & \underline{22.73} & \underline{0.7106} & \underline{0.1692} & \underline{22.69} & \underline{0.7118} & \underline{0.1652} \\
\textbf{Ours}                       & \textbf{23.38} & \textbf{0.7389} & \textbf{0.1718} & \textbf{23.09} & \textbf{0.6989} & \textbf{0.1700} & \textbf{23.41} & \textbf{0.7300} & \textbf{0.1652} & \textbf{23.45} & \textbf{0.7312} & \textbf{0.1641} \\
\bottomrule
\end{tabular}
}
\label{tab:noncircular_compare}
\vspace{4mm}

\begin{minipage}[t]{0.58\textwidth}
\centering
\captionof{table}{\textbf{Quantitative comparison of circular bokeh rendering} on synthetic MultiLens and real-world EBB400 dataset.}
\resizebox{\linewidth}{!}{
\begin{tabular}{l ccc ccc}
\toprule
\multirow{2}{*}{\textbf{Method}} &
\multicolumn{3}{c}{\textbf{MultiLens-Circle}} &
\multicolumn{3}{c}{\textbf{EBB400}~\cite{ignatov2020rendering}} \\
\cmidrule(lr){2-4} \cmidrule(lr){5-7}
 & PSNR $\uparrow$ & SSIM $\uparrow$ & LPIPS $\downarrow$
 & PSNR $\uparrow$ & SSIM $\uparrow$ & LPIPS $\downarrow$ \\
\midrule
MPIB~\cite{peng2022mpib}      & 21.22 & 0.6797 & 0.4284 & 23.08 & 0.7855 & 0.3032 \\
BokehMe~\cite{peng2022bokehme} & 21.34 & 0.6749 & 0.4407 & 23.15 & 0.7880 & 0.2984 \\
Dr.Bokeh~\cite{sheng2024dr}    & 19.83 & 0.6397 & 0.4629 & 22.27 & 0.7501 & 0.3510 \\
BokehDiff~\cite{zhu2025bokehdiff} & \underline{22.70} & \textbf{0.7642} & \underline{0.2368} & \underline{23.31} & \textbf{0.8029} & \underline{0.2558} \\
\textbf{Ours} & \textbf{23.33} & \underline{0.7387} & \textbf{0.1743} & \textbf{23.58} & \underline{0.7892} & \textbf{0.2398} \\
\bottomrule
\end{tabular}
}
\label{tab:circular_compare_two_datasets}
\end{minipage}
\hfill
\begin{minipage}[t]{0.40\textwidth}
\centering
\captionof{table}{\textbf{User study results} indicate that users prefer ours regarding visual quality.}
\resizebox{\linewidth}{!}{
\begin{tabular}{lc}
\toprule
Comparison & Human Preference \\
\midrule
Ours vs. SuperEdit~\cite{li2025superedit} & \textbf{76.8}\% / 23.2\% \\
Ours vs. UltraEdit~\cite{zhao2024ultraedit} & \textbf{88.4}\% / 11.6\% \\
Ours vs. SuperEdit*~\cite{li2025superedit} &  \textbf{60.8}\% / 39.2\%\\
Ours vs. UltraEdit*~\cite{zhao2024ultraedit} & \textbf{76.5}\% / 23.5\% \\
\bottomrule
\end{tabular}
}
\label{tab:user_study}
\end{minipage}

\end{table*}

\subsubsection{Quantitative Comparison.}

As shown in Table~\ref{tab:circular_compare_two_datasets}, for circular bokeh rendering, LensStyle consistently outperforms all neural approaches on both synthetic \textbf{MultiLens} and the real-world\textbf{ EBB400}~\cite{ignatov2020rendering} dataset, achieving the highest PSNR and SSIM scores and the lowest LPIPS scores. Although the SSIM is lower than that of diffusion-based BokehDiff, our method achieves better performance in terms of LPIPS, which is more insensitive to blur. 
More importantly, quantitative comparisons of other lens styles are reported in Table~\ref{tab:noncircular_compare}.
Existing diffusion-based image editing models (e.g., SuperEdit~\cite{li2025superedit}, UltraEdit~\cite{zhao2024ultraedit}) fail to produce physically accurate and stylistically consistent effects, as they lack 
knowledge
of lens geometry, vignetting, and diffraction. 
After being fine-tuned on our MultiLens dataset, their performance improves but still fall short of ours due to the lack of lens-aware design.
In contrast, our model achieves a significant improvement of over 1.1--1.8 dB in PSNR and 15--25\% relative reduction in LPIPS, indicating better perceptual fidelity and aesthetic consistency.
This validates the importance of the proposed Dual Path Controller and its continuous-discrete conditioning mechanism in capturing both the optical parameters and semantic aspects of real lens styles.

\begin{figure*}[!t]
	\centering
	\includegraphics[width=\linewidth]{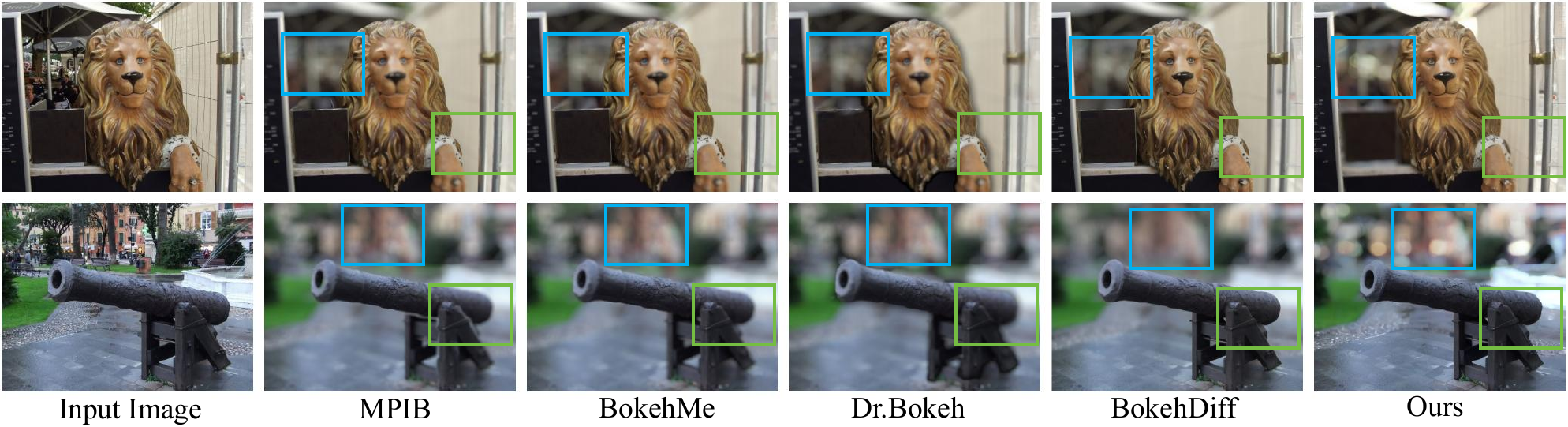}
	\caption{\textbf{Qualitative Comparison against bokeh rendering methods for standard circle bokeh.} All baselines take a depth map predicted by Depth Anything V2~\cite{yang2024depth} as input. As a result, when depth map is inaccurate, the focused region of output can be blurred by mistake (green box area). Moreover, ours produces more aesthetically pleasing circle bokeh (blue box area).
	}
	\label{fig:circle}
\end{figure*}

\begin{figure*}[!t]
	\centering
	\includegraphics[width=\linewidth]{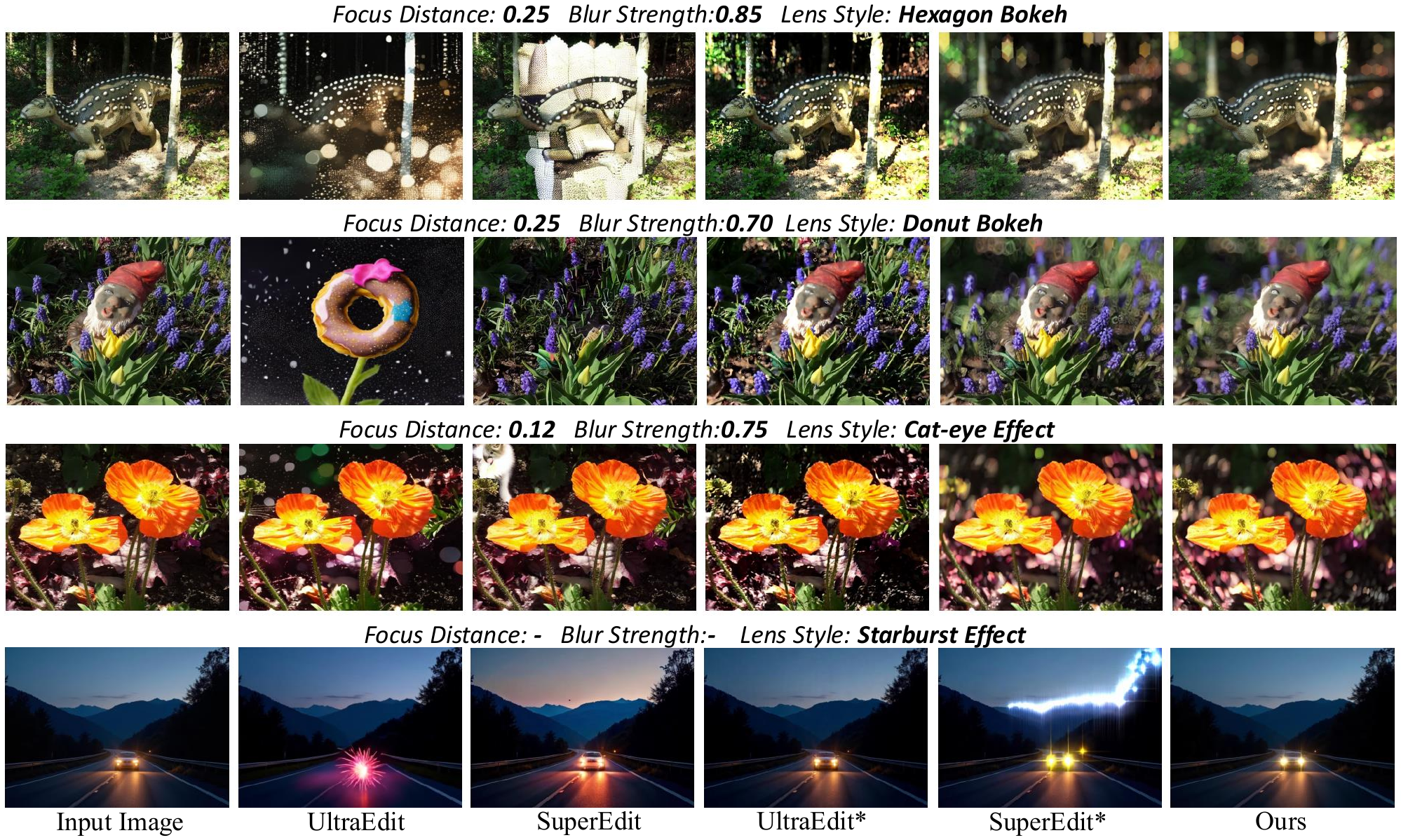}
	\caption{\textbf{Qualitative Comparison against diffusion-based image editing methods.} UltraEdit and SuperEdit lack lens style awareness, resulting in outputs that do not match the specified lens style. After fine-tuning on our MultiLens dataset, these models begin to capture certain lens style characteristics. However, thanks to the proposed Dual Path Controller with its disentangled continuous-discrete conditioning mechanism, our method achieves the most consistent and visually superior results.  
	}
	\label{fig:bokeh_others}
\end{figure*}
\subsubsection{Qualitative Comparison.}
For standard circle bokeh rendering, Figure~\ref{fig:circle} presents visual comparisons among different bokeh rendering methods.
All baselines share the same depth map predicted by Depth Anything V2~\cite{yang2024depth}, while ours is in a depth-free manner. As a result, when the depth input is inaccurate in some scenarios, the focused region can be mistakenly blurred, such as the green box areas in Figure~\ref{fig:circle}. Furthermore, benefiting from the designed Lens Style Controller, ours produces more physically consistent and visually pleasing circular bokeh, such as blue box areas.


Figure~\ref{fig:bokeh_others} compares other lens styles with diffusion-based image editing methods. Since UltraEdit and SuperEdit lack lens-aware related knowledge, they produce outputs that are semantically inaccurate with respect to the editing instructions. After fine-tuned on our MultiLens dataset, these models begin to capture certain lens style characteristics. However, they either produce less physically consistent bokeh patterns (e.g., the hexagon bokeh in the first row of Figure~\ref{fig:bokeh_others}) or over-produces starbursts (e.g., the sky in the fourth row of Figure~\ref{fig:bokeh_others}).
Our method still achieves the most consistent and visually superior results, benefiting from the proposed Dual Path Controller with its disentangled continuous modulation and discrete cross-attention mechanism.


       
        
     
\subsection{User Study}
\label{subsec:userstudy}

Since no publicly available benchmark provides real paired images across diverse lens styles for perceptual evaluation, we conduct a user study to assess aesthetic quality and style fidelity.
We collect 32 all-in-focus images from Unsplash~\cite{Unsplash}, a photography platform containing real-world captured images. 
The selected images cover diverse content categories, including portraits, animals, indoor and outdoor scenes, as well as daytime and nighttime settings. 
Each image is rendered with different methods to produce corresponding lens-style outputs under matched style prompts and optical parameters.

We perform a pairwise A/B comparison protocol. 
For each trial, participants are shown two rendered images generated from the same input image and style configuration. 
One image is always produced by our method, while the other is randomly selected from a competing approach. 
The presentation order is randomized to avoid bias.
Participants are asked to choose the one that (1) exhibits stronger aesthetic quality, (2) better reflects the intended lens style, and (3) maintains higher consistency with the input image content. 

\begin{figure*}[!t]
	\centering
    \includegraphics[width=\textwidth]{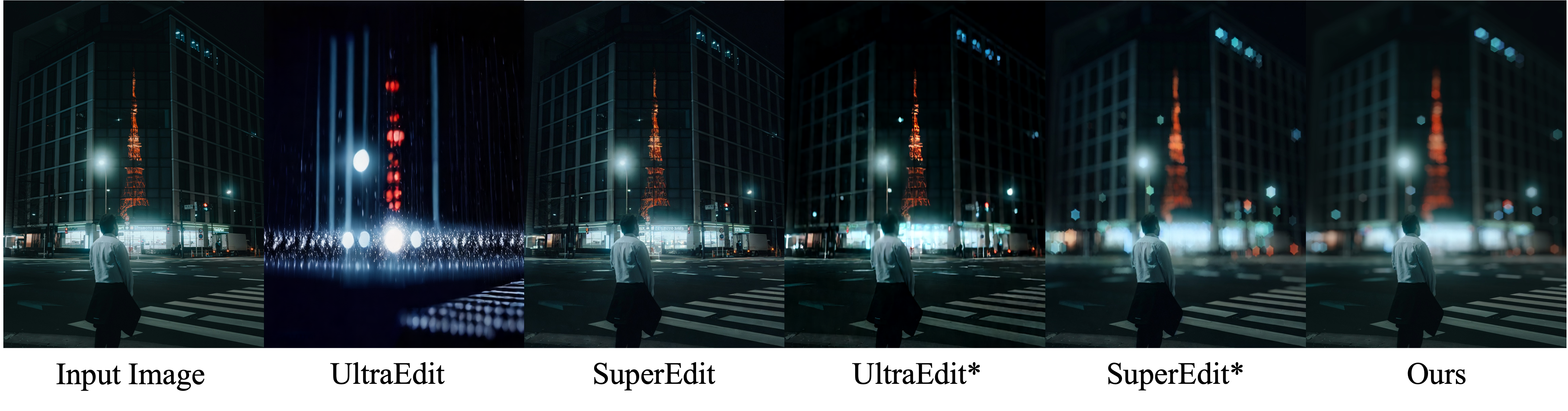}
	\caption{\textbf{User study visual comparisons} under focus distance $0.70$ and blur intensity $0.80$ with hexagonal aperture style. Our method produces hexagonal bokeh patterns that are geometrically consistent and physically plausible, whereas SuperEdit*~\cite{li2025superedit} generates hexagons with overly sharp corner transitions.
	}
	\label{fig:userstudy}
\end{figure*}

A total of 58 participants took part in the study.
As summarized in Table~\ref{tab:user_study} and illustrated in Figure~\ref{fig:userstudy}, 
our method is consistently preferred over competing approaches. 
The results demonstrate that LensStyle achieves superior perceptual realism, stylistic fidelity, and structural consistency, validating its effectiveness in multi-style lens effect rendering.


\begin{table*}[t]
\centering
\caption{\textbf{Ablation study on the MultiLens dataset.}
We evaluate the contribution of each core component in LensStyle.
}
\begin{tabular}{lcccccc}
\toprule
\textbf{Variant} 
& \textbf{Flow} 
& \textbf{COPM} 
& \textbf{DLSC} 
& \textbf{PSNR} $\uparrow$ 
& \textbf{SSIM} $\uparrow$ 
& \textbf{LPIPS} $\downarrow$ \\
\midrule
w/o Flow (Direct Regression) &  & \checkmark & \checkmark & 21.87 & 0.6822 & 0.2433 \\
w/o OPM                      & \checkmark &  & \checkmark & 22.14 & 0.7034 & 0.2126 \\
w/o SCA                      & \checkmark & \checkmark &  & 22.06 & 0.6951 & 0.2195 \\
\midrule
\textbf{Full Model}          & \checkmark & \checkmark & \checkmark 
& \textbf{23.38} & \textbf{0.7389} & \textbf{0.1718} \\
\bottomrule
\end{tabular}
\label{tab:ablation}
\end{table*}
    

\subsection{Ablation Study}
\label{subsec:ablation}

We conduct ablation experiments to analyze the contribution of each key component in LensStyle, including the continuous optical modulation, discrete style conditioning, and flow-based transport.




\subsubsection{Continuous Optical Parameter Modulation(COPM).}

To validate the effectiveness of adaptive feature modulation, 
we replace the $\gamma(\mathbf{p}), \beta(\mathbf{p})$ modulation with simple concatenation of parameter embeddings to the input features.

This modification significantly weakens smooth parameter interpolation.
In particular, blur magnitude transitions become less monotonic when sweeping focus distance or blur strength.
These results indicate that affine feature modulation provides more stable and interpretable continuous control than naive conditioning.

\subsubsection{Discrete Lens Style Cross-Attention(DLSC).}

We further remove the cross-attention branch and inject style embeddings only through global concatenation.
This leads to weaker aperture-specific characteristics, especially for complex styles such as donut bokeh and starburst diffraction.
This demonstrates that cross-attention effectively governs spatially structured lens-style semantics.

\subsubsection{Flow-Based Transport vs. Direct Regression.}

To examine the impact of the flow formulation, 
we replace the flow matching objective with a direct latent regression loss:
\[
\mathcal{L}_{reg} = \| \hat{z}_L - z_L \|_2^2.
\]
This variant produces less stable structural preservation and more over-smoothed outputs.
Flow-based transport better maintains input content consistency while allowing large stylistic transformations, suggesting that learning a velocity field provides a more effective mapping between the two domains.

Overall, these ablation results validate the necessity of each design component in achieving controllable and realistic multi-style lens effect rendering.

\section{Conclusion and Future Work}
\label{sec:conclusion}
In this work, we introduced LensStyle, a unified framework for controllable multi-style lens effect rendering.
Unlike conventional defocus rendering approaches that focus primarily on blur magnitude, LensStyle explicitly models lens aesthetics by jointly learning continuous optical parameter modulation and discrete lens-style conditioning within a flow-based generative architecture.
By leveraging a physically grounded optical forward model for dataset synthesis, our method learns depth-consistent blur transitions while remaining depth-free during inference. 
The proposed Dual Path Controller disentangles continuous focus and blur control from aperture-specific stylistic characteristics, enabling interpretable and flexible manipulation of lens effects in a single model.
Extensive quantitative evaluations and user studies demonstrate that LensStyle achieves improved realism, style fidelity, and controllability compared to existing lens rendering and image editing approaches.

\noindent \textbf{Limitation and future work.} While LensStyle effectively models diverse lens aesthetics and achieves controllable lens effect rendering, it cannot simulate bokeh effects of different DSLRs such as Fuji, Sony, Nikon and Canon. Future work will extend our model to lens of different brands.



%
%
\bibliographystyle{splncs04}
\bibliography{main}
\end{document}